\documentclass[12pt]{article}

\usepackage{latexsym,enumerate}
\usepackage{amsmath,amsthm,amsopn,amstext,amscd,amsfonts,amssymb,euler, amsbsy}
\usepackage{amssymb}

\def \<{\langle}
\def\>{\rangle}

\def \R{\Bbb R}
\def \Rd{\Bbb R^d}

\def \K{\mathcal K}

\def \dt{\delta}

\def \R{\Bbb R}

\def \fee{\varphi}
\def \e{\epsilon}

\def \s{\sigma}

\def\eps{{\varepsilon}}

\newcommand\eref[1]{(\ref{#1})}

\newtheorem{proposition}{Proposition}
\newtheorem{lemma}{Lemma}
\newtheorem{theorem}{Theorem}

\newtheorem{corollary}{Corollary}
\newtheorem{remark}{Remark}

\newenvironment{Proof}{\noindent{\bf Proof:}\quad}{\endproof}

\begin{document}

\title{\bf Optimal exponential bounds on the accuracy of classification}


%\author[label1]{G. Kerkyacharian
%, D. Picard
%, V. Temlyakov, A. Tsybakov
%}
%\author[label1]{
%G. Kerkyacharian
%D. Picard
%V. Temlyakov, A. Tsybakov
%}
%\address[label1]{Universit� Paris-Diderot, CNRS-LPMA, 175 rue du Chevaleret, 75013 %Paris, France}
%\author[label2]{
%G. Kerkyacharian, D. Picard,
%V. Temlyakov
%A. Tsybakov
%}
%\address[label2]{Department of Mathematics,
%University of South Carolina, Columbia, SC29208-USA}
%\author[label3]{
%G. Kerkyacharian, D. Picard,
%V. Temlyakov,
%A.B. Tsybakov
%}
%\address[label3]{CREST-ENSAE, 3 av. Pierre Larousse, 92240 Malakoff,
%France}
%\author[label4]{V. Koltchinskii}
%\address[label4]{Department of Mathematics, Georgia
%Institute of Technology}

%\author{{G. Kerkyacharian, D. Picard}
%\\
%\footnotesize Universit\'e Paris-Diderot, CNRS-LPMA
%\\
%\footnotesize 175 rue du Chevaleret,
%75013 Paris, France
%\\
%{V. Temlyakov}
%\\
%\footnotesize Department of Mathematics,
%University of South Carolina
%\\
%\footnotesize Columbia, SC 29208, USA
%\\
%{A.B. Tsybakov}
%\\
%\footnotesize CREST-ENSAE, 3 av. Pierre Larousse, 92240 Malakoff,
%France
%\\
%{V. Koltchinskii}
%\thanks{Partially supported by NSF Grants DMS-0906880 and CCF-0808863}
%\\ \footnotesize School of Mathematics, Georgia Institute of Technology
%\\ \footnotesize Atlanta, GA 30332-0160, USA
%%\\ \footnotesize vlad@math.gatech.edu
%}
%

\author{{N.I. Pentacaput}
\\
\footnotesize {\it n.i.pentacaput@gmail.com} }

\maketitle

\begin{abstract}
Consider a standard binary classification problem, in which $(X,Y)$
is a random couple in ${\cal X}\times \{0,1\}$ and the training data
consists of $n$ i.i.d. copies of $(X,Y).$ Given a binary classifier
$f:{\cal X}\mapsto \{0,1\},$ the generalization error of $f$ is
defined by $R(f)={\mathbb P}\{Y\neq f(X)\}$. Its minimum $R^*$ over
all binary classifiers $f$ is called the Bayes risk and is attained
at a Bayes classifier. The performance of any binary classifier
$\hat f_n$ based on the training data is characterized by the excess
risk $R(\hat f_n)-R^*$. We study Bahadur's type exponential bounds
on the following minimax accuracy confidence function based on the
excess risk:
$$
{AC}_n({\cal M},\lambda) =\inf_{\hat f_n}\sup_{P\in {\cal M}}\mathbb P\left(
R(\hat{f}_n)-R^*\ge \lambda\right), \lambda \in [0,1],
$$
where the supremum is taken over all distributions $P$ of $(X,Y)$ from
a given class of distributions ${\cal M}$ and the infimum is over all
binary classifiers $\hat f_n$ based on the training data. We study how
this quantity depends on the complexity of the class of distributions
${\cal M}$ characterized by exponents of entropies of the class of regression
functions or of the class of Bayes classifiers corresponding to the distributions
from ${\cal M}.$ We also study its dependence on margin parameters of the
classification problem.
In particular, we show that, in the case when
${\cal X}=[0,1]^d$ and ${\cal M}$ is the class all distributions
satisfying the margin condition with exponent $\alpha>0$ and
such that the regression function $\eta$ belongs to a given H\"older
class of smoothness $\beta>0,$
$$
-\frac{\log AC_n(\mathcal{M}, \lambda)}{n} \asymp
\lambda^{\frac{2+\alpha}{1+\alpha}},\ \lambda \in [Dn^{-\frac{1+\alpha}{2+\alpha+d/\beta}},\lambda_0]
$$
for some constants $D,\lambda_0>0.$

%In this paper we measure the accuracy of classifiers using Bahadur's type exponential %bounds.
%Similar results have been obtained in the regression setting in
% \cite{DKPT}:
%Namely, it is proved there that there exist $\lambda_n^-,\; \lambda_n^+$, with %$\lambda_n^-\asymp \lambda_n^+\asymp n^{-r/(1+2r)}$, and constants $\delta_0,\; %C_1,\; c_1,\; C_2,\; c_2$ such that
%\begin{align*}
%\inf_{\hat \xi_n\in\mathbb{S}_n}\sup_{P\in\mathcal{M}(\Theta) }\mathbb P\left(
%\|\hat \xi_n-\xi\|_{\mu_X}\ge \lambda\right)&\ge \delta_0,\qquad \forall\; \lambda\le %\lambda_n^-,\\
%C_1 e^{-c_1n\lambda^2}\le \inf_{\hat
%\xi_n\in\mathbb{S}_n}\sup_{P\in\mathcal{M}(\Theta) }\mathbb P\left(
%\|\hat \xi_n-\xi\|_{\mu_X}\ge \lambda\right)&\le C_2
%e^{-c_2n\lambda^2},\qquad \forall\; \lambda\ge \lambda_n^+.
%\end{align*}
%where $r$ is an index describing the entropy of the parameter set.
%We prove here comparable results in the classification setting, for classes of %probabilities indexed by an entropy condition (index $r$) as well as a margin %condition (associated to a coefficient $\alpha$).
%The exponential rate is then of the form $\exp\{-c'n %\lambda^{\frac{2+\alpha}{1+\alpha}}  \},$ and the critical values $\lambda_n^-$ and $ %\lambda_n^+$ are of the form $n^{-\frac{1+\alpha}{2+\alpha(1+r)}}$.
\end{abstract}

{\footnotesize {\bf AMS classification:} 62G08, 62G07, 62H05, 68T10}

{\footnotesize {\bf Key words and phrases:} statistical learning,
classification, fast rates, optimal rate of convergence, excess
risk, margin condition, Bahadur efficiency}

%\begin{keyword}

%AMS classification 62G08, 62G07, 62H05, 68T10. learning theory, classification,  exponential rates.
%\end{keyword}

%\end{frontmatter}

% \linenumbers

\section{Introduction}
Let $(\mathcal{X},{\cal A})$ be a measurable space. We consider a
random variable $(X,Y)$ in $\mathcal{X}\times\{0,1\}$ with
probability distribution denoted by $P$. Denote by $\mu_X$ the
marginal distribution of $X$ in $\mathcal{X}$ and by
$$
\eta(x)\triangleq\eta_P(x)\triangleq{P}(Y=1|X=x)={E}(Y|X=x)
$$
the conditional probability of $Y=1$ given $X=x$, which is also the
regression function of $Y$ on $X$. Assume that we have $n$ i.i.d.
observations of the pair $(X,Y)$ denoted by ${\mathcal
D}_n=((X_i,Y_i))_{i=1,\ldots,n}$. The aim is to predict the output
label $Y$ for any input $X$ in $\mathcal{X}$ from the observations
${\mathcal D}_n$.

We recall some standard facts of classification theory. A
\textit{prediction rule} is a measurable function $f\; :\;
\mathcal{X}\longmapsto\{0,1\}$. To any prediction rule we associate
the \textit{classification error} (probability of
misclassification):
\[
R(f) \triangleq {P}\bigl(Y\neq f(X)\bigr).
\]
It is well
known (see, e.g., Devroye \textit{et al.} \cite{Devroye96})
that
\[
\min_{f\; :\;\mathcal{X}\longmapsto\{0,1\}} R(f) = R(f^*)\triangleq
R^{*},
\]
where the prediction rule $f^*$, called the \textit{Bayes rule}, is defined by
\[
f^*(x)\triangleq f^*_P(x)\triangleq {{I}}_{\{\eta(x)\ge
1/2\}},\qquad \forall x\in\mathcal{X},
\]
where ${{I}}_{A}$ denotes the indicator function of $A$.
The minimal risk $R^*$ is called the \textit{Bayes risk}. A
\textit{classifier} is a function, $\hat{f}_n=\hat{f}_n(X,{\mathcal
D}_n)$, measurable with respect to ${\mathcal D}_n$ and $X$ with
values in $\{0,1\}$, that assigns to the sample ${\mathcal D}_n$ a
prediction rule $\hat{f}_n(\cdot,{\mathcal D}_n)\; :\;
\mathcal{X}\longmapsto\{0,1\}$. A key characteristic of $\hat{f}_n$
is its risk $\mathbb{E}[R(\hat{f}_n)]$, where
\[
R(\hat{f}_n) \triangleq {P}\bigl(Y\neq\hat {f}_n(X) |{\mathcal
D}_n\bigr).
\]
The aim of statistical learning is to construct a classifier
$\hat{f}_n$ such that $R(\hat{f}_n)$ is as close to $R^*$ as
possible. The accuracy of a classifier $\hat{f}_n$ is usually
measured by the quantity $%$\mathcal E(\hat {f}_n)=
\mathbb{E}[R(\hat {f}_n) - R^*]$ called the (expected)
\textit{excess risk} of $\hat{f}_n$, where the expectation
$\mathbb{E}$ is taken with respect to the distribution of ${\mathcal
D}_n$. We say that the classifier $\hat{f}_n$ learns with the
convergence rate $\psi(n)$, if there exists an absolute constant
$C>0$ such that for any integer $n$, $\mathbb{E}[R(\hat{f}_n) -
R^*]\leq C\psi(n)$.

Given a convergence rate, Theorem 7.2 of Devroye  \textit{et al.}
\cite{Devroye96} shows that no classifier can learn with this rate
for {\it all} underlying probability distributions~$P$. To achieve
some rates of convergence, we need to restrict the class of possible
distributions $P.$ For instance, Yang \cite{Yang99} provides
examples of classifiers learning with a given convergence rate under
complexity assumptions expressed via the smoothness properties of
the regression function $\eta$. Under complexity assumptions alone,
no matter how strong they are, the rates cannot be faster than
$n^{-1/2}$ (cf. Devroye \textit{et al.} \cite{Devroye96}).
Nevertheless, they can be as fast as $n^{-1}$ if we add a control on
the behavior of the regression function $\eta$ at the level $1/2$
(the distance $|\eta(\cdot)-1/2|$ is sometimes called the {margin}).
This behavior is usually characterized by the following condition
introduced in~\cite{tsyb04}.

%%%%%%%%%%%

\smallskip

 {\bf Margin  condition.} \textit{The probability distribution $P$ on
the space $\mathcal{X}\times\{0,1 \}$ satisfies the Margin condition
with exponent $0<\alpha<\infty$ if there exists $C_M>0$ such that}
\begin{equation}\label{2.15}
\mu_X\big(0<|\eta(X)-1/2|\leq t\big)\leq C_M
t^{\alpha},\qquad\forall0\leq t <1.
\end{equation}

\smallskip

Equivalently, one can assume that (\ref{2.15}) holds only for $t\in
[0,t_0]$ for some $t_0\in [0,1).$ This would imply (\ref{2.15}) for
all $t\in [0,1)$ (with a larger value of $C_M$). In this form,
(\ref{2.15}) makes sense also for $\alpha=+\infty,$ it is
interpreted as $\mu_X\big(0<|\eta(X)-1/2|\leq t_0\big)=0,$ and it
was used, e.g., in~\cite{MassartNedelec}. Another equivalent form of
margin condition (\ref{2.15}) is discussed in the next section (see
(\ref{2.12})) and it is characterized by the margin parameter
$\kappa=(1+\alpha)/\alpha$ ($\kappa=1$ for $\alpha=+\infty$). Under
the margin condition, \textit{fast rates}, that is, rates faster
than $n^{-1/2}$ can be obtained for different classifiers, cf.
Tsybakov  \cite{tsyb04}, Blanchard \textit{et al.} \cite{BLV03},
Bartlett \textit{et al.} \cite{BJM03}, Tsybakov and van de Geer
\cite{tsy_vdg}, Koltchinskii \cite{Koltch}, Massart and N\'ed\'elec
\cite{MassartNedelec}, Audibert and Tsybakov \cite{Audtsyb}, Scovel
and Steinwart \cite{ScSt} among others.

In this paper, we will study the closeness of $R(\hat{f}_n)$  to
$R^*$ in a more refined way. Our measure of performance  is inspired
by the Bahadur efficiency of estimation procedures but on the
difference from the classical Bahadur approach (cf., e.g.,
\cite{ibrhasm}) we obtain non-asymptotic results.

For a classifier $\hat{f}_n$ and for a tolerance $\lambda>0,$
define {\it the accuracy confidence function} (or, shortly, the
$AC$-function):
\begin{equation}{AC}_n(\hat{f}_n,\lambda) =\mathbb P\left(
R(\hat{f}_n)-R^*\ge \lambda\right). \label{accuracy}\end{equation}
Here $\mathbb{P}$ denotes the probability distribution of the
observed sample ${\mathcal D}_n$. Note that
${AC}_n(\hat{f}_n,\lambda)=0$ for $\lambda>1$ since $0\le R(f)\le 1$
for all classifiers~$f$. Moreover, $R(\hat{f}_n)-R^*\le 1/2$ for all
interesting classifiers $\hat{f}_n$. Indeed, it makes no sense to
deal with the probabilities of error $R(\hat{f}_n)$ greater than
$1/2$ (note that $R(\hat{f}_n)=1/2$ is achieved when $\hat{f}_n$ is
the simple random guess classifier). Therefore, without loss of
generality we can consider only $\lambda\le 1/2$. In fact,we will
sometimes use a slightly stronger restriction $\lambda\le \lambda_0$
for some $\lambda_0<1/2$ independent of $n$.

It is intuitively clear that if the tolerance is low ($\lambda$
under some critical value $\lambda_n$), the probability
${AC}_n(\hat{f}_n,\lambda)$ is kept larger than some fixed level. On
the opposite, for $\lambda\geq \lambda_n$, the quality of the
procedure $\hat{f}_n$ can be characterized by the rate of
convergence of ${AC}_n(\hat{f}_n,\lambda)$ towards zero as
$n\to\infty$. Observe that evaluating the critical value $\lambda_n$
yields, as a consequence, bounds and the associated rates for the
excess risk $\mathbb E R(\hat{f}_n)-R^*$, which is a commonly used
measure of performance.

For a class $\mathcal{M}$ of probability measures $P$, we define the
minimax $AC$-function
 \begin{equation}\label{ACfn}
 AC_n(\mathcal{M},\lambda)\triangleq\inf_{\hat f_n\in\mathbb{S}_n}\sup_{P\in\mathcal{M} }\mathbb P\left(
R(\hat{f}_n)-R^*\ge \lambda\right), \end{equation} where
$\mathbb{S}_n$ is the set of all classifiers. We will consider
classes ${\mathcal M}={\mathcal M}(r , \alpha)$ defined by the
following conditions:
\begin{itemize}
\item[(a)] A margin assumption with exponent $\alpha$.
\item[(b)] A complexity assumption expressed in terms of the rate of
decay $r>0$ of an $\eps$-entropy.
\end {itemize}
%(see section \ref{lower} for details)
The main results of this paper can be summarized as follows. Fix $r
, \alpha>0$ and set $\lambda_n= D n^{-\frac{1+\alpha}{2+\alpha+r'}}$
where $D>0$, and $r'=r'(\alpha,r)>0$ is a function of $\alpha$ and
$r$ depending on the type of the imposed complexity assumptions.
Then, we have an upper bound: There exist positive constants $C,\;
c$ such that, for all classes ${\mathcal M}={\mathcal M}(r ,
\alpha)$ satisfying the above two  conditions,
\begin{equation}\label{quote00}
AC_n(\mathcal{M}, \lambda)\le C\exp\{-cn
\lambda^{\frac{2+\alpha}{1+\alpha}}\}, \quad \forall \; \lambda\ge
\lambda_n.
\end{equation}
Furthermore, we prove the corresponding lower bound: there exists a
class ${\mathcal M}$  satisfying the same conditions (a) and (b)
such that
\begin{align}
AC_n(\mathcal{M},\lambda)&\ge p_0,\qquad 0<\lambda\le \lambda_n^-\asymp \lambda_n,\\
AC_n(\mathcal{M}, \lambda)&\ge C' \exp\{-c'n
\lambda^{\frac{2+\alpha}{1+\alpha}}\}, \quad  \lambda_n\asymp \lambda_n^+\le
\lambda\le \lambda_0\, \label{quote02}
\end{align}
for some positive constants $p_0,\; C',\; c'$ and $0<\lambda_0<1/2$
depending only on $C_M$ and $\alpha$. Thus, we quantify the critical
level phenomenon discussed above and we derive the exact exponential
rate $\exp\{-cn \lambda^{\frac{2+\alpha}{1+\alpha}}\}$ for minimax
$AC$-function over the critical level. In particular, this implies
the following bounds on the minimax $AC$-function in the case when
${\cal X}=[0,1]^d$ and ${\cal M}$ is the class all distributions
satisfying the margin condition with exponent $\alpha>0$ and such
that the regression function $\eta$ belongs to a H\"older class of
smoothness $\beta>0$ (see Section \ref{sec:lower:holder}):
\begin{eqnarray}
&&
\nonumber
AC_n(\mathcal{M},\lambda)\ge p_0,\
0<\lambda\le D_1 n^{-\frac{1+\alpha}{2+\alpha+d/\beta}},
\\
&&
\nonumber
C'\exp\{-c'n
\lambda^{\frac{2+\alpha}{1+\alpha}}\}\le
AC_n(\mathcal{M}, \lambda) \le
C\exp\{-cn
\lambda^{\frac{2+\alpha}{1+\alpha}}\},
\\
&&
\nonumber
D_2 n^{-\frac{1+\alpha}{2+\alpha+d/\beta}}\le
\lambda\le \lambda_0.
\end{eqnarray}

As an immediate consequence of (\ref{quote00}) -- (\ref{quote02}) we
get the minimax rate for the excess risk:
\begin{equation}
\inf_{\hat f_n\in\mathbb{S}_n}\sup_{P\in\mathcal{M} } \big[ \mathbb
E R(\hat{f}_n)-R^* \big] \asymp n^{-\frac{1+\alpha}{2+\alpha+r'}}\,
\end{equation}
for appropriate classes $\mathcal{M}$, which implies the results previously obtained in Tsybakov \cite{tsyb04} and Audibert and Tsybakov \cite{Audtsyb}.

It is interesting to compare (\ref{quote00}) -- (\ref{quote02}) to the results for the regression problem in a similar setting (see DeVore \textit{et al.} \cite{DKPT} and Temlyakov \cite{Temapprox}) since
there are similarities and differences.
Let us quote these former results: suppose, in a supervised learning setting, that we observe   $n$ i.i.d. observations of the pair $(X,Y)$, but here $Y$ is valued in $[-M,M]$ instead of $\{0,1\}$ and we want to estimate
$$\xi(x)={\mathbb E}(Y|X=x).$$
Let $\hat \xi_n(x)$ denote an estimator of $\xi(x)$ and consider the loss
$$\|\hat \xi_n-\xi\|_{\mathbb{L}_2(\mu_X)}.$$
Here and in what follows, $\|\cdot\|_{\mathbb{L}_p(\mu_X)}$,
$p\ge1$, denotes the $\mathbb{L}_p(\mu_X)$-norm with respect to the
measure $\mu_X$ on $\mathcal{X}$.  In this context,
$AC_n(\mathcal{M},\lambda)$ denotes  the quantity
$$\inf_{\hat \xi_n}\sup_{P\in\mathcal{M} }\mathbb P\left(
\|\hat \xi_n-\xi\|_{\mathbb{L}_2(\mu_X)}\ge \lambda\right).$$ It is
proved in  \cite{DKPT} and \cite{Temapprox} that if
$\mathcal{M}=\mathcal{M}(\Theta,\mu_X)$ is the set of probability
measures having $\mu_X$ as marginal distribution and such that $\xi$
belongs to the set $\Theta$, and the  entropy numbers of $\Theta$
with respect to $\mathbb{L}_2(\mu_X)$ are of order $n^{-r}$ (see
\cite{DKPT} and \cite{Temapprox} for details), then there exist
$\lambda_n^-,\; \lambda_n^+$, with $\lambda_n^-\asymp
\lambda_n^+\asymp n^{-r/(1+2r)}$, and constants $\delta_0,\; C_1,\;
c_1,\; C_2,\; c_2$ such that
\begin{align}\label{quote1}
AC_n(\mathcal{M}(\Theta,\mu_X) ,\lambda)&\ge \delta_0,\qquad \forall\; \lambda\le \lambda_n^-,\\
C_1 e^{-c_1n\lambda^2}\le AC_n(\mathcal{M}(\Theta,\mu_X),
\lambda)&\le C_2 e^{-c_2n\lambda^2},\qquad \forall\; \lambda\ge
\lambda_n^+. \label{quote2}
\end{align}
These inequalities describe accurately the behavior of the minimax
$AC$-function for classes $\mathcal{M}(\Theta,\mu_X)$ with any
marginal distribution $\mu_X$. The same inequalities hold for the
following quantity
$$
\sup_{\mu_X}AC_n(\mathcal{M}(\Theta,\mu_X),\lambda).
$$
Our results for the classification problem are somewhat weaker than
the above results for the regression problem. In Sections
\ref{sec:upper:reg} and \ref{sec:upper:Bayes}, we prove the upper
bounds for the corresponding classes in the case of any marginal
distribution $\mu_X$ such that the Margin assumption holds. This is
analogous to what was obtained for the regression problem. However,
in Section \ref{sec:lower:holder}, we only prove the matching lower
bounds for a special marginal distribution $\mu_X$. Thus we obtain
an accurate description of the behavior of the supremum over
marginal distributions $ \sup_{\mu_X}AC_n(\mathcal{M},\lambda) $ and
not of the individual $AC$-functions for each marginal distribution
$\mu_X$.

The similarity  of the results in the two different settings is that
there is a regime of exponential concentration, which holds for any
$\lambda$ greater than a critical level. This critical level, which
is also the minimax rate, depends on the complexity of the class
characterized by $r$. We can also observe that the exponents in the
bounds (${\frac{2+\alpha}{1+\alpha}} $ in classification, $2$ in
regression) do not depend on the complexity parameter $r$.

The differences lie in two facts since the margin condition is
entering the game at two levels. The first one is the critical value
itself, $n^{-\frac{1+\alpha}{2+\alpha+r'}}$. Note that here $\alpha$
is appearing in a favorable way (the larger it is, the better the
rate). This is intuitively clear since larger $\alpha$ correspond to
sharper decision boundaries.

The second place where a difference occurs is the rate in  the
exponent $ \lambda^{\frac{2+\alpha}{1+\alpha}} $ compared to
$\lambda ^2$ in a regression setting. The margin condition
influences the rate ${\frac{2+\alpha}{1+\alpha}} $, and this time
again in a favorable way with respect to $\alpha$ (the rate improves
as $\alpha$ grows).  For $\alpha\to 0$, that is, when there is no
margin condition we approach the same rate as in regression.

%As detailed in section \ref{lower} the lower bound is obtained as a consequence of a more general %inequality.

\section{Properties related to the Margin condition}

In this section, we discuss some facts related to the Margin
condition. We first recall that it can be equivalently defined in
the following way, cf. \cite{tsyb04}.
\begin{proposition}\label{prop2.1} A probability measure $P$
satisfies the Margin condition \eref{2.15}
if and only if there exists a positive  constant $c_M$ such that,
for any Borel set $G\subset {\mathcal X}$,
\begin{equation}\label{2.12}
\int_G |2\eta(x)-1|\mu_X(dx) \ge c_M\mu_X(G)^{\varkappa},
\end{equation}
where $\varkappa=(1+\alpha)/\alpha$.
\end{proposition}
\begin{Proof} Let $G$ be given. Clearly, it suffices to assume that
$\mu_X(G) >0$. Choose $t$ from the equation $\mu_X(G)=2C_Mt^\alpha$.
Then by the Margin condition
$$
\mu_X(G\setminus\{0<|\eta(X)-1/2|\le t\}) \ge \mu_X(G) -C_Mt^\alpha
\geq C_M t^{\alpha}.
$$
Therefore,
\begin{eqnarray}\label{margin_equiv}
\int_G|2\eta(x)-1|\mu_X(dx)&\ge &
2\int_{G\setminus\{x:0<|\eta(x)-1/2|\le t\}} t\mu_X(dx) \\ \nonumber
&\ge& 2C_Mt^{\alpha+1} = (2C_M)^{-1/\alpha}\mu_X(G)^{1+1/\alpha}.
\end{eqnarray} Conversely,
assume that for some $\varkappa>1$ inequality (\ref{2.12}) holds for
any Borel set $G$. Take $G=\{x:0<|\eta(x)-1/2|\le t\}$. Then
(\ref{2.12}) yields
\begin{eqnarray*}
\mu_X(0<|\eta(X)-1/2|\le t)&\le&
\left(c_M^{-1}\int_{0<|\eta(x)-1/2|\le t} |2\eta(x)-1|\mu_X(dx) \right)^{1/\varkappa}\\
&\le& \big(2c_M^{-1}t \,\mu_X (0<|\eta(X)-1/2|\le t)
\big)^{1/\varkappa}.
\end{eqnarray*}
Solving this inequality with respect to $\mu_X(0<|\eta(X)-1/2|\le
t)$ we obtain the Margin condition \eref{2.15}. \qed
\end{Proof}

\begin{remark}\label{remark:1} The constant $C_M$ in Margin
condition \eref{2.15} satisfies
$$
C_M\ge 1/2.
$$
\end{remark}
\begin{Proof} By \eqref{margin_equiv} we have that (\ref{2.12}) holds with
constant $c_M=(2C_M)^{-1/\alpha}$. Using this and the fact that
$0\le \eta(x)\le 1$ we get $\mu_X(G)\ge c_M \mu_X(G)^\varkappa =
(2C_M)^{-1/\alpha} \mu_X(G)^{\frac{1+\alpha}{\alpha}}$ for all
$G\subset {\mathcal X}$. Thus, $2C_M \ge \mu_X(G)$, and since this
holds for all $G$ and $\mu_X$ is a probability measure we get the
result.
\end{Proof}

\begin{remark}\label{remark:1''}
The statement of Proposition \ref{prop2.1} also holds with
$\kappa=1$ for the case $\alpha=+\infty$, which is understood as
discussed after the definition of Margin condition~\eref{2.15}.
\end{remark}

We now state an easy consequence of Proposition \ref{prop2.1}.

\begin{lemma}\label{lem2.3} If the probability measure
$P$ satisfies the Margin condition \eref{2.15}, then for any
prediction rule $f$,
$$
R(f)-R^* \ge
(2C_M)^{-1/\alpha}\|f-f^*_P\|_{L_1(\mu_X)}^{\frac{1+\alpha}{\alpha}}.
$$
Analogously, if the probability measure $P$ satisfies the Margin
condition \eref{2.12} with some $\varkappa\ge 1$, then for any
prediction rule $f$,
$$
R(f)-R^* \ge c_M\|f-f^*_P\|_{L_1(\mu_X)}^{\varkappa}.
$$
\end{lemma}
\begin{Proof} Note that, for any
prediction rule $f$,
\begin{equation}
 R(f)-R^* = \int_{D_P(f)}|2\eta(x)-1| \mu_X (dx), \label{1.1}
\end{equation}
where $D_P(f)\triangleq\{x:f^*_P(x)\neq f(x)\}$. By \eqref{margin_equiv} we have that (\ref{2.12}) holds with
constant $c_M=(2C_M)^{-1/\alpha}$. Thus, the result follows from
\eref{2.12} and the obvious relation
%\begin{align}
% R(\hat f_n)-R^* &= \int_{D_P(\hat f_n)}|2\eta(x)-1| d\mu_X  \label{2.13}\\
%&\ge c\mu_X(D_P(\hat f_n))^{\frac{1+\alpha}{\alpha}}.\label{2.14}
%\end{align}
%It remains to notice that
$$
\mu_X(D_P(f))=\|f-f^*_P\|_{L_1(\mu_X)}. %=\int_X|\hat f_n-f^*_P|d\mu_X =
%\int_{D_P(\hat f_n)}|\hat f_n-f^*_P|d\mu_X
%\mu_X(D_P(f)).
$$
\end{Proof}
Finally, we will use the following property.

%%%%%%

\begin{proposition}\label{prop2.2}
For any Borel function $\bar \eta:{\mathcal X}\to [0,1]$ and any
distribution $P$ of $(X,Y)$ satisfying the Margin
condition~\eref{2.15}, we have
$$
\|f_{\bar \eta}-f^*_P\|_{L_1(\mu_X)}\le 2C_M\|\bar
\eta-\eta_P\|_{L_\infty(\mu_X)}^{\alpha}\,
$$
where $f_{\bar \eta}(x)={{I}}_{\{\bar\eta(x)\ge 1/2\}}.$
\end{proposition}
\begin{Proof} By Lemma 5.1 in \cite{Audtsyb},
\begin{equation}\label{at51}
R(f_{\bar \eta})-R^*\le 2C_M \|\bar
\eta-\eta_P\|_{L_\infty(\mu_X)}^{1+\alpha}. \end{equation} This and
Lemma \ref{lem2.3} yield the result.
\end{Proof}

\begin{corollary} \label{cor1} Let $\mathcal P$ be a class of joint distributions
of $(X,Y)$ satisfying the Margin condition \eref{2.15} and all
having the same marginal $\mu_X$. Then, for any pair $P, \bar P\in
\mathcal P$ with the corresponding regression functions $\eta, \bar
\eta$ and decision rules $f_{\eta}(x)={{I}}_{\{\eta(x)\ge 1/2\}}$,
$f_{\bar \eta}(x)={{I}}_{\{\bar\eta(x)\ge 1/2\}}$, we have
$$
\|f_{\bar \eta}-f_{\eta}\|_{L_1(\mu_X)}\le 2C_M\|\bar
\eta-\eta\|_{L_\infty(\mu_X)}^{\alpha}\,.
$$
\end{corollary}

%Finally, we introduce a slightly modified margin condition.
%%%%%%%%%%

%\noindent
%{\bf Bracket Margin condition} \textit{The
%probability distribution $P$ on the space $\mathcal{X}\times\{0,1
%\}$
%satisfies the Bracket Margin condition  with $1\leq\varkappa<+\infty$ if there exist constants
%$c>0$ and $c'>0$ such that, for any Borel set $G\subset {\mathcal X}$,
%}
%\begin{equation}\label{bm_cond}
%c \mu_X(G)^{\varkappa}\le
% \int_G |2\eta(x)-1|d\mu_X \le c' \mu_X(G)^{\varkappa}.
%\end{equation}
%In view of Proposition \ref{prop2.1}, the Bracket Margin condition
%is more restrictive than the Margin condition \eref{2.15}: Bracket
%Margin condition with $\kappa=(1+\alpha)/\alpha$ implies the Margin
%condition with parameter $\alpha$.

\bigskip

%%%%%%%%%%%%%%%%%%%%%%%%%

\section{Upper bound under complexity assumption on the regression
function}\label{sec:upper:reg}

%%%%%%%%%%%%%%%%%%

%For a sake of simplicity, in this section we identify a classifier
%with a set $G$: $\hat f_n=I_{G}$. Let $G^*=\{x:\eta(x)\ge1/2\}$ be
%the set corresponding to the Bayes classifier. Hence we write the
%risk of a classifier in the form $R(G)$. Define $d_\triangle
%(G,G^*)\stackrel {\mathrm{def}}{=}\mu_X(G\triangle G^*)$ and
%$$
%d(G,G^*)\stackrel {\mathrm{def}}{=} \int_{G\triangle
%G^*}|2\eta(x)-1| \mu_X(dx) = R(G)-R(G^*),
%$$
%where $G\triangle G^*$ denotes the symmetric difference of the sets
%$G$ and $G^*$.

%In this section, we assume that $\mathcal X$ is a metric space and

In this section, we prove an upper bound of the form \eqref{quote00}
for a class of probability distributions $P$, for which the
complexity assumption (b) (cf. the Introduction) is expressed in
terms of the entropy of the class of underlying regression functions
$\eta_P$.

For $g: \mathcal X\to \R$, define the sup-norm $\|g\|_\infty=
\sup_{x\in \mathcal X}|g(x)|$.

Fix some positive constants $r,\alpha, C_M, B.$ Let ${\mathcal M}(r,
\alpha)={\mathcal M}(r, \alpha,C_M, B)$ be any set of joint
distributions $P$ of $(X,Y)$ satisfying the following two
conditions.
%Denote by ${\mathcal M}(\mu_X,r, \alpha)$ the
%set of joint distributions $P$ of $(X,Y)$ with common marginal
%distribution $\mu_X$ satisfying the following two conditions.
{\it
\begin{itemize}
\item[(i)] The Margin condition \eref{2.15} with exponent $\alpha$
and constant $C_M$.
\item[(ii)] The regression function $\eta=\eta_P$ belongs to
a known class of functions~${\mathcal U}$, which admits the
$\eps$-entropy bound
\begin{equation}\label{U1}
%\forall \ \mu_X: \quad
\mathcal H(\eps,{\mathcal U},\|\cdot\|_\infty)\le B\eps^{-r}, \
\forall \e>0.
\end{equation}
Here, the $\eps$-entropy $\mathcal H(\eps,{\mathcal
U},\|\cdot\|_{\infty})$ is defined as the
%$\log_2$
natural logarithm of the minimal number of $\eps$-balls in the
$\|\cdot\|_{\infty}$ norm needed for covering~${\mathcal U}$.
\end{itemize}
}
For any prediction rule $f$, we define the empirical risk
$$
R_n(f)=\frac1{n}\sum_{i=1}^n I_{\{f(X_i)\ne Y_i\}}\,.
$$
We consider the classifier $\hat f_{n,1}(x) =I_{\{\hat \eta_n(x)\ge
1/2\}},$ where
$$
\hat \eta_n = {\rm argmin}_{\eta'\in
{\mathcal N}_{\varepsilon}} R_n(f_{\eta'}).
$$
Here $f_{\eta'}(x) =I_{\{\eta'(x)\ge 1/2\}}$ and ${\mathcal
N}_{\varepsilon}$ denotes a minimal ${\varepsilon}$-net on
${\mathcal U}$ in the $\|\cdot\|_{\infty}$ norm, i.e., ${\mathcal
N}_{\varepsilon}$ is the minimal subset of ${\mathcal U}$ such that
the union of $\eps$-balls in the $\|\cdot\|_{\infty}$ norm centered
at the elements of~${\mathcal N}_{\varepsilon}$ covers~${\mathcal
U}$.

%%%%%%%%%%%%%%%%%%%%%%%%%%%%%%%%
\begin{theorem}\label{theoup}
Let $r,\alpha, C_M, B$ be finite positive constants. Set
${\varepsilon}={\varepsilon}_n=n^{-\frac{1}{2+\alpha+r}}$.
%be defined as in \eref{def:eps_n}.
Then there exist positive constants $c$ and $c'$ depending only on
$r
, \alpha, C_M, B$ such that%, for any $\mu_X$,
$$\sup_{P\in {\mathcal M}(r , \alpha)} \mathbb P\{
R(\hat f_{n,1})-R(f^*_P)\ge \lambda\}\le 2\exp\{-cn
\lambda^{\frac{2+\alpha}{1+\alpha}} \}
$$
for  $\lambda\ge c' n^{-\frac{1+\alpha}{2+\alpha+r}}$.% where $d=???
%\max\{2c_M^2, (2Bc_3)^{\frac{2+\alpha}{1+\alpha}}\}$ with
%$c_3=???32(c_M^1)^{-\frac{\alpha}{1+\alpha}}+ 16/3.$
\end{theorem}
This theorem has an immediate consequence in terms of
$AC$-functions.
\begin{corollary}
 There exist $d>0$, $c>0$ such that for   $\lambda_n = d
 n^{-\frac{1+\alpha}{2+\alpha+r}}$ we have
 \begin{equation}\label{quote12}
%\forall \ \mu_X: \quad
AC_n({\mathcal M}(r , \alpha) ,\lambda)\le 2
e^{-cn\lambda^{\frac{2+\alpha}{1+\alpha}}},\qquad \forall\;
\lambda\ge \lambda_n.
\end{equation}
\end{corollary}

\noindent{\sc Proof of Theorem \ref{theoup}.} Set $d(\eta')\triangleq
R(f_\eta')-R(f^*_P)$. Let $\bar\eta\in {\mathcal N}_{\varepsilon}$ be
such that $\|\bar\eta-\eta_P\|_\infty \le \varepsilon$. Using
(\ref{at51}) we get
\begin{equation}\label{u1}
d(\bar \eta)=R(f_{\bar\eta}) - R^* \le 2C_M
\|\bar\eta-\eta_P\|_\infty^{1+\alpha} \le 2C_M
\varepsilon^{1+\alpha} \le \lambda/2
\end{equation}
for any $\lambda\ge 4C_M n^{-\frac{1+\alpha}{2+\alpha+r}}$. Define a
set of functions ${\mathcal G}_{\varepsilon}=\{\eta'\in {\mathcal
N}_{\varepsilon}: d(\eta') \ge \lambda \}$, and introduce the centered empirical
increments
$$
{\mathcal Z}_n(\eta')=(R_n(f_{\eta'})-R_n(f^*_P)) -
(R(f_{\eta'})-R(f^*_P)).
$$
Then
\begin{eqnarray*}
\mathbb P(R(\hat f_{n,1})-R(f^*_P) \ge\lambda) &\le& \mathbb
P(\exists \eta'\in {\mathcal G}_{\varepsilon}:
R_n(f_\eta')-R_n(f_{\bar \eta})\le 0)\\
&\le&\sum_{\eta'\in {\mathcal G}_{\varepsilon}}\mathbb
P(d(\eta')+{\mathcal Z}_n(\eta')-d(\bar\eta) -{\mathcal
Z}_n(\bar\eta)\le 0).
\end{eqnarray*}
Note that for any $\eta'\in {\mathcal G}_{\varepsilon}$ we have
$$
d(\eta') - d(\bar\eta) \ge d(\eta')/2 \ge \lambda/2.
$$
Using this remark and (\ref{U1}) we find
\begin{eqnarray}\label{u2}
\mathbb P(R(\hat f_{n,1})-R(f^*_P)  \ge\lambda) &\le& \sum_{\eta'\in
{\mathcal G}_{\varepsilon}} \mathbb P({\mathcal Z}_n(\eta')\le -d(\eta')/4)\\
&&+ \mathbb P({\mathcal Z}_n(\bar\eta)\ge \lambda/4)\nonumber
\\
&\le& \exp (B\varepsilon^{-r}) \max_{\eta'\in {\mathcal
G}_{\varepsilon}}\mathbb P({\mathcal Z}_n(\eta')\le -d(\eta')/4)
\nonumber\\ &&+ \mathbb P({\mathcal Z}_n(\bar\eta)\ge
\lambda/4).\nonumber
\end{eqnarray}
Now, ${\mathcal Z}_n(\eta')=\frac1{n}\sum_{i=1}^n \xi_i(\eta'),$ where
$$\xi_i(\eta')=I_{\{f_{\eta'}(X_i)\ne Y_i\}}-I_{\{f^*_P(X_i)\ne Y_i\}}-
{\mathbb E}\Big(I_{\{f_{\eta'}(X_i)\ne Y_i\}}-I_{\{f^*_P(X_i)\ne
Y_i\}}\Big)\,.
$$
Clearly, $|\xi_i(\eta')|\le 2$ and, by Lemma \ref{lem2.3},
\begin{eqnarray*}
{\mathbb E}(\xi_i(\eta')^2) &\le&
{\mathbb E}\Big(\big[I_{\{f_{\eta'}(X_i)\ne Y_i\}}-I_{\{f^*_P(X_i)\ne Y_i\}}\big]^2\Big)\\
&=&\|f_{\eta'} - f_P^*\|_{L_1(\mu_X)}\\
&\le&\left[(2C_M)^{1/\alpha}(R(f_{\eta'})-R(f^*_P))\right]
^{\frac{\alpha}{1+\alpha}}
\\
&=& (2C_M)^{\frac1{1+\alpha}}d^{\frac{\alpha}{1+\alpha}}(\eta').
\end{eqnarray*}
Therefore, we can apply Bernstein's inequality to get
\begin{align*}
\mathbb P({\mathcal Z}_n(\eta')\le -d(\eta')/4) &\le
\exp\left(-\frac{nd^2(\eta')/16}{2((2C_M)^{\frac1{1+\alpha}}
d^{\frac{\alpha}{1+\alpha}}(\eta')
+ d(\eta')/3)}\right)\\
&\le
\exp\left(-\frac{nd^2(\eta')}{c_1'd^{\frac{\alpha}{1+\alpha}}(\eta')}\right)
\end{align*}
where $c_1'=2((2C_M)^{\frac1{1+\alpha}}+1/3)$ and we used that
$d(\eta')\le d^{\frac{\alpha}{1+\alpha}}(\eta')$ since $d(\eta')\le 1$.
Thus, for any $\eta'\in {\mathcal G}_{\varepsilon}$ we obtain
$$
\mathbb P({\mathcal Z}_n(\eta')\le
-d(\eta')/4)\le\exp\left(-n\lambda^{\frac{2+\alpha}{1+\alpha}}/c_1'\right).
$$
As a consequence,
\begin{eqnarray}\nonumber
\exp (B\varepsilon^{-r}) \max_{\eta'\in {\mathcal G}_{\varepsilon}}
\mathbb P({\mathcal Z}_n(\eta')\le -d(\eta')/4) &\le&
\exp(Bn^{\frac{r\alpha}{2+\alpha+r}}-
n\lambda^{\frac{2+\alpha}{1+\alpha}}/c_1')\\
&\le& \exp(-n\lambda^{\frac{2+\alpha}{1+\alpha}}/2c_1')\label{u3}
\end{eqnarray}
where we used that $\lambda\ge c' n^{-\frac{1+\alpha}{2+\alpha+r}}$
for some large enough $c'>0$. Another application of Bernstein's
inequality and (\ref{u1}) yields
\begin{align*}
\mathbb P({\mathcal Z}_n(\bar\eta)\ge \lambda/4) &\le
\exp\left(-\frac{n\lambda^2/16}{2((2C_M)^{\frac1{1+\alpha}}
d^{\frac{\alpha}{1+\alpha}}(\bar\eta)
+ \lambda/3)}\right) \\
&\le
\exp\left(-\frac{n\lambda^2}{c_1'(\lambda^{\frac{\alpha}{1+\alpha}}
+ \lambda)}\right)\,.
\end{align*}
For $\lambda\le1$ the last inequality implies
$$
\mathbb P({\mathcal Z}_n(\bar\eta)\ge \lambda/4) \le
\exp\left(-\frac{n\lambda^{\frac{2+\alpha}{1+\alpha}}}{2c_1'}\right).
$$
This, together with (\ref{u2}) and (\ref{u3}), yields result of
the theorem for $\lambda\le1$. If $\lambda>1$ it holds trivially
since $d(\eta')\le 1$ for all $\eta'$.

%\medskip

%%%%%%%%%%%%%%%%%%%%%%%%%%%%%%%

\medskip

\section{Upper bound under complexity assumption on the Bayes
classifier}\label{sec:upper:Bayes}

\medskip

In this section, we prove a result analogous to those of
Section~\ref{sec:upper:reg} when the complexity assumption (b) (cf.
the Introduction) is expressed in terms of the entropy of the class
of underlying Bayes classifiers $f^*_P$ rather than of that of
regression functions $\eta_P$.

%In this section, we derive an exponential inequality for the
%excess risk of classification rules based on empirical risk
%minimization (ERM) over a set of binary classifiers that
%contains the Bayes rule.

First, introduce some definitions. Let ${\cal F}$ be a class of
measurable functions from a measurable space $(S,{\cal A}_S, \mu)$
into $[0,1]$. Here $\mu$ is a $\sigma$-finite measure. For $1\le
q\le \infty$, and $\eps>0$, let $N_{[\ ]}(\eps, {\cal F},
\|\cdot\|_{L_q(\mu)})$ denote the $L_q(\mu)$-bracketing numbers of
${\cal F}.$ That is, $N_{[\ ]}(\eps, {\cal F},
\|\cdot\|_{L_q(\mu)})$ is the minimal number $N$ of functional
brackets
$$
[f_j^-,f_j^+]\triangleq\{g: f_j^-\leq g\leq f_j^+\},\ j=1,\dots, N,
$$
such that
$$
{\cal F}\subset \bigcup_{j=1}^N [f_j^-,f_j^+]\ \ {\rm and}\
\ \|f_j^{+}-f_j^-\|_{L_q(\mu)}\leq \eps,\ j=1,\dots, N.
$$
The bracketing $\eps$-entropy of ${\cal F}$ in the
$\|\cdot\|_{L_q(\mu)}$-norm is defined by
$$\mathcal H_{[\ ]}(\eps, {\cal F}, \|\cdot\|_{L_q(\mu)})\triangleq\log N_{[\ ]}
(\eps, {\cal F}, \|\cdot\|_{L_q(\mu)}).$$

We will consider a class of probability distributions $P$ of
$(X,Y)$ characterized by the complexity of the corresponding
Bayes classifiers. Specifically, fix some $\rho\in
(0,1),0<\alpha\le \infty, c_M>0, c_\mu>0, B'>0$, and let %${\mathcal M}^*(\rho, \alpha)$
%the set of joint distributions $P$ of $(X,Y)$ satisfying the
%following conditions.
%%Denote by
${\mathcal M}^*(\rho, \alpha)= {\mathcal M}^*(\rho, \alpha, c_M,
c_\mu, B')$ be any set of joint distributions $P$ of $(X,Y)$
satisfying the following conditions. {\it
\begin{itemize}
\item[(i)] The marginal distribution $\mu_X$ of $X$ is absolutely
continuous with respect to a $\sigma$-finite measure $\mu$ on
$({\cal X},{\cal A})$, and $(d\mu_X/d\mu )(x) \le c_\mu$ for
$\mu$-almost all $x\in {\cal X}$.
\item[(ii)] The Margin condition \eref{2.12} with exponent
$\varkappa=(1+\alpha)/\alpha$ and constant~$c_M$ is satisfied (we
adopt the convention that $\varkappa=1$ corresponds to
$\alpha=\infty$).
\item[(iii)] The Bayes classifier $f^*_P$ belongs to
a known class of prediction rules~${\mathcal F}$ satisfying the
bracketing entropy bound
\begin{equation}\label{U1_2}
%\forall \ \mu_X: \quad
\mathcal H_{[\ ]}(\eps,{\mathcal F}, \|\cdot\|_{L_1(\mu)})\le
B'\eps^{-\rho}, \ \forall \eps>0.
\end{equation}
\end{itemize}
}

The results below still hold in this slightly more general situation.

We consider a classifier $\hat f_{n,2}$ that minimizes the empirical
risk over the class ${\cal F}:$
$$
\hat f_{n,2} \triangleq {\rm argmin}_{f\in {\cal F}} R_n(f).
$$
The main result of this section is that for $\hat f_{n,2}$ we have
the following exponential upper bound.

%%%%%%%%%%%%
\begin{theorem}\label{th:upper:2}
Let $\rho\in (0,1),0<\alpha\le \infty$, and let $c_M, c_\mu, B'$ be
positive constants. Then there exist positive constants $c$ and $c'$
depending only on $\rho, \alpha, c_M, c_\mu, B'$ such that
$$\sup_{P\in {\mathcal M}^*(\rho, \alpha)} \mathbb P\{
R(\hat f_{n,2})-R(f^*_P)\ge \lambda\}\le e \exp\{-cn
\lambda^{\frac{2+\alpha}{1+\alpha}} \}
$$
for  $\lambda\ge c' n^{-\frac{1+\alpha}{2+\alpha(1+\rho)}}$. Here we
adopt the convention that $\frac{2+\alpha}{1+\alpha}=1$, and
$\frac{1+\alpha}{2+\alpha(1+\rho)}= \frac1{1+\rho}$ for
$\alpha=\infty$.
\end{theorem}

We deduce Theorem \ref{th:upper:2} from the following fact that we
state here as a proposition.

%%%%%%%%%%%%%%%%%%%%%%%%%%%%%%%%
\begin{proposition}
\label{upper_class_ERM} There exists a constant $C_*>0$
depending only on $\rho, \alpha, C_M$ such that, for all $t>0,$
$$
\sup_{P\in {\mathcal M}^*(\rho, \alpha)}{\mathbb P}\biggl\{ R(\hat
f_{n,2})-R(f^*_P)\geq C_*
\biggl[n^{-\frac{\varkappa}{2\varkappa-1+\rho}}\vee
\biggl(\frac{t}{n}\biggr)^{\frac{\varkappa}{2\varkappa-1}} \biggr]
\biggr\}\leq e^{1-t}.
$$
\end{proposition}

It is easy to see that Theorem \ref{th:upper:2} follows from this
proposition by taking $t=c n \lambda^{\frac{2+\alpha}{1+\alpha}}$
with $\lambda \geq c' n^{-\frac{1+\alpha}{2+\alpha(1+\rho)}}$ for
some constants $c,c'>0,$ and using that
$\varkappa=\frac{1+\alpha}{\alpha}$ if $\alpha<\infty$.

Proposition \ref{upper_class_ERM} will be derived from a general
excess risk bound in abstract empirical risk minimization
(\cite{koltch_11}, Theorem 4.3). We will state this result here for
completeness. To this end, we need to introduce some notation. Let
${\cal G}$ be a class of measurable functions from a probability
space $(S,{\cal A}_S, P)$ into $[0,1]$ and let $Z_1,\dots, Z_n$ be
i.i.d. copies of an observation $Z$ sampled from $P.$ For any
probability measure $P$ and any $g\in {\cal G}$, introduce the
following notation for the expectation:
$$
Pg=\int_S gdP.
$$
Denote by $P_n$ the empirical measure based on $(Z_1,\dots, Z_n)$,
and consider the minimizer of empirical risk
$$
\hat g_n \triangleq{\rm argmin}_{g\in {\cal G}}P_n g.
$$
For a function $g\in {\cal G},$ define the excess risk
$$
{\cal E}_P(g)\triangleq Pg-\inf_{g'\in {\cal G}}Pg'.
$$
The set
$$
{\cal F}_P(\delta)\triangleq\{g\in {\cal G}: \, {\cal E}_P(g)\leq
\delta\}
$$
is called the $\delta$-minimal set. The size of such a set will
be controlled in terms of its $L_2(P)$-diameter
$$
D(\delta)\triangleq\sup_{g,g'\in {\cal
F}_P(\delta)}\|g-g'\|_{L_2(P)}
$$
and also in terms of the following ``localized empirical
complexity'':
$$
\phi_n(\delta)\triangleq{\mathbb E}\sup_{g,g'\in {\cal
F}_P(\delta)}|(P_n-P)(g-g')|.
$$
We will use these complexity measures to construct an upper
confidence bound on the excess risk ${\cal E}_P(\hat f_{n,2}).$ For
a function $\psi:{\mathbb R}_+\mapsto {\mathbb R}_+,$ define
$$
\psi^{\flat}(\delta)\triangleq\sup_{\sigma \geq
\delta}\frac{\psi(\sigma)}{\sigma}.
$$
Let
$$
V_n^t(\delta)\triangleq4\biggl[\phi_n^{\flat}(\delta)+\sqrt{(D^2)^{\flat}(\delta)\frac{t}{n\delta}}+
\frac{t}{n\delta}\biggr],\ \delta>0, t>0,
$$
and define
$$
\sigma_n^t\triangleq\inf\{\sigma: V_n^t(\sigma)\leq 1\}.
$$

The following result is the first bound of Theorem 4.3 in
\cite{koltch_11}.

\begin{proposition}
\label{abstract_ERM} For all $t>0,$
$$
{\mathbb P}\{{\cal E}_P(\hat f_{n,2})>\sigma_n^t\}\leq e^{1-t}.
$$
\end{proposition}

In addition to this, we will use the well-known inequality for the
expected sup-norm of the empirical process in terms of bracketing
entropy, see Theorem 2.14.2 in \cite{vVW96}. %van der Vaart and Wellner
%(1996).
More precisely, we will need the following simplified version of
that result.

\begin{lemma}
\label{bracket_bound} Let ${\cal T}$ be a class  of functions from
$S$ into $[0,1]$ such that $\|g\|_{L_2(P)}\leq a$ for all $g\in
{\cal T}.$ Assume that $H_{[\ ]}(a, {\cal
T},\|\cdot\|_{L_2(P)})+1\le a^2 n$. Then
$$
{\mathbb E}\sup_{g\in {\cal T}}|P_n g -P g|\leq \frac{\bar C}{\sqrt{n}} \int_0^{a}
\left(H_{[\ ]}(\eps, {\cal T},\|\cdot\|_{L_2(P)})+1\right)^{1/2}
d\eps,
$$
where $\bar C>0$ is a universal constant.
\end{lemma}

{\bf Proof of Proposition \ref{upper_class_ERM}}. Note that, if
$t>n,$ then $(\frac{t}{n})^{\varkappa/(2\varkappa-1)}> 1,$ and the
result holds trivially with $C_*=1$ since $R(\hat
f_{n,2})-R(f^*_P)\leq 1.$ Thus, it is enough to consider the case
$t\leq n.$

Let $S={\cal X}\times \{0,1\}$ and $P$ be the distribution of
$Z=(X,Y)$. We will apply Proposition~\ref{abstract_ERM} to the class
${\cal G}\triangleq\{g_f: \, g_f(x,y)=I_{\{y \ne f(x)\}}, \ f\in {\cal
F}\}$. Then, clearly, $Pg_f=R(f)$ and ${\cal E}_P (g_f)= R(f)-R(f^*_P)$
for $g_f(x,y)=I_{\{y \ne f(x)\}},$ which implies that
$$
{\cal F}_P(\delta)=\{g_f: f\in {\cal F},\ R(f)-R(f^*_P)\leq \delta\}.
$$
We also have $\|g_{f_1}-g_{f_2}\|_{L_2(P)}^2=\|f_1-f_2\|_{L_1(\mu_X)}.$
Thus, it follows from Lemma~\ref{lem2.3} that, for all
$g_f\in {\cal G}$,
$$
{\cal E}_P (g_f)\geq c_M \|g_f-g_{f_{P}^{\ast}}\|_{L_2(P)}^{2\varkappa}
$$
and we get a bound on the $L_2(P)$-diameter of the $\delta$-minimal
set ${\cal F}_P(\delta):$ with some constant ${\bar c}_1>0$
\begin{eqnarray}\label{ddelta}
&&D(\delta)\leq {\bar c}_1 \delta^{1/(2\varkappa)}.
\end{eqnarray}
To bound the function $\phi_n(\delta),$ we will apply
Lemma \ref{bracket_bound} to the class ${\cal T}={\cal F}_P(\delta)$
with $a=1$.
%$$\sigma \leq D(\delta)\leq {\bar c}_1 \delta^{1/(2\varkappa)}.$$
Note that
\begin{eqnarray*}
 H_{[\ ]}(\eps, {\cal
F}_P(\delta),\|\cdot\|_{L_2(P)})&\leq& 2H_{[\ ]}(\eps/2,{\cal
G},\|\cdot\|_{L_2(P)})\\
&\leq& 2 H_{[\ ]}(\eps^2/4, {\cal F},\|\cdot\|_{L_1(\mu_X)})
\\ &\le& 2
H_{[\ ]}(\eps^2/(4c_\mu), {\cal F},\|\cdot\|_{L_1(\mu)}) .
\end{eqnarray*}
Using (\ref{U1_2}) we easily get from Lemma \ref{bracket_bound}
that, with some constants ${\bar c}_2, {\bar c}_3>0$,
$$
\phi_n(\delta)\leq {\bar c}_2 \delta^{\frac{1-\rho}{2\varkappa}}
n^{-1/2},\ \  \delta \geq {\bar c}_3 n^{-\frac{\varkappa}{1+\rho}},
$$
which implies that, with some constant ${\bar c}_4>0$,
$$
\phi_n(\delta)\leq {\bar c}_4
\max(\delta^{\frac{1-\rho}{2\varkappa}} n^{-1/2},
n^{-\frac{1}{1+\rho}}),\delta >0.
$$
This and (\ref{ddelta}) lead to the following bound on the function
$V_n^t(\delta)$:
$$
V_n^t(\delta)\leq {\bar c}_5
\biggl[\delta^{\frac{1-\rho}{2\varkappa}-1}n^{-1/2}\vee
\delta^{-1}n^{-\frac{1}{1+\rho}}+\delta^{\frac{1}{2\varkappa}-1}
\sqrt{\frac{t}{n}}+
\delta^{-1}\frac{t}{n}\biggr]
$$
that holds with some constant ${\bar c}_5.$ Thus, we end up with a
bound on $\sigma_n^t:$
\begin{equation}
\label{sigm_bd} \sigma_n^{t}\leq {\bar c}_6
\biggl[n^{-\frac{\varkappa}{2\varkappa-1+\rho}}\vee
n^{-\frac{1}{1+\rho}}\vee
\biggl(\frac{t}{n}\biggr)^{\varkappa/(2\varkappa-1)} \vee
\frac{t}{n} \biggr].
\end{equation}
Note that, for $\varkappa\geq 1,$ $\rho< 1$ and $t\leq n,$ we have
$$
n^{-\varkappa/(2\varkappa-1+\rho)}\geq n^{-1/(1+\rho)}\ \ {\rm and}\
\ \left(\frac{t}{n}\right)^{\varkappa/(2\varkappa-1)} \geq
\frac{t}{n}.
$$
Therefore, (\ref{sigm_bd}) can be simplified as follows:
\begin{equation}
\label{sigm_bd_1} \sigma_n^{t}\leq {\bar c}_7
\biggl[n^{-\frac{\varkappa}{2\varkappa-1+\rho}}+
\biggl(\frac{t}{n}\biggr)^{\varkappa/(2\varkappa-1)} \biggr],
\end{equation}
and the result immediately follows from Proposition
\ref{abstract_ERM}.  \qed

%%%%%%%%%

Note that Theorem~\ref{th:upper:2} remains valid if we drop
condition (i) and replace (iii) by the following more general
condition:
{\it
\begin{itemize}
\item[(iii')] The Bayes classifier $f^*_P$ belongs to
a known class of prediction rules~${\mathcal F}$ satisfying the
bracketing entropy bound
\begin{equation}\label{U1_2''}
%\forall \ \mu_X: \quad
\mathcal H_{[\ ]}(\eps,{\mathcal F}, \|\cdot\|_{L_1(\mu_X)})\le
B'\eps^{-\rho}, \ \forall \eps>0.
\end{equation}
\end{itemize}
}
Condition (iii') is, in fact, an assumption on both~${\mathcal F}$
and the class of possible marginal densities $\mu_X$. The reason why
we have introduced conditions (i) and (iii) instead of (iii') is
that they are easily interpretable. Indeed, in this way we decouple
assumptions on ${\mathcal F}$ and $\mu_X$. The case that is even
easier corresponds to considering a subclass of ${\mathcal
M}^*(\rho, \alpha)$ composed of measures $P\in {\mathcal M}^*(\rho,
\alpha)$ with the same marginal $\mu_X$. Then again we only need to
assume (ii) and (iii') but now (iii') should hold for one fixed
measure $\mu_X$ and not simultaneously for a set of possible
marginal measures.
%Moreover, we have the following inclusion.
%%
%%
%%
%\begin{remark}\label{remark:2} Let $0<\alpha<\infty$. Let ${\mathcal M}(\mu_X,r, \alpha)$
%be a subset of ${\mathcal M}(r, \alpha)$ composed of measures with
%the same fixed marginal $\mu_X$. Then
%$$
%{\mathcal M}(\mu_X,\rho\alpha, \alpha) \subseteq {\mathcal
%M}^*(\mu_X,\rho, \alpha).
%$$
%\end{remark}
%\begin{proof}
%\end{proof}

We finish this section by a comparison of Theorems~\ref{theoup}
and~\ref{th:upper:2}. They differ in imposing entropy assumptions on
different objects, regression function $\eta_P$ and Bayes classifier
$f^*_P$ respectively. Also, in Theorem~\ref{theoup} the complexity
is measured by the usual entropy for the sup-norm, whereas in
Theorem~\ref{th:upper:2} it is done in terms of the bracketing
entropy for the $L_1$-norm. Note that for many classes the
bracketing and the usual $\eps$-entropies behave similarly, so that
the relationship between the corresponding rates of decay $r$ in
(\ref{U1}) and $\rho$ in (\ref{U1_2}) is only determined by the
relationship between the sup-norm of the regression function $\eta$
and the $L_1$-norm on the induced Bayes classifier. In this respect,
Corollary~\ref{cor1} is insightful suggesting the correspondence
$\rho=r/\alpha$. In the next section, we will see that such a
correspondence exactly holds when the regression function $\eta$
belongs to a H\"older class. Finally, note that the ranges of the
margin and complexity parameters as well as the assumptions on the
measure $\mu_X$ in Theorems~\ref{theoup} and~\ref{th:upper:2} are
somewhat different. Namely, Theorem~\ref{theoup} holds under no
additional assumption on $\mu_X$ except for the Margin condition and
covers classes with high complexity (all $r>0$ are allowed).
Theorem~\ref{th:upper:2} needs a relatively mild additional
assumption (i) on $\mu_X$ and restricts the complexity by the
condition $\rho<1.$ On the other hand, Theorem~\ref{th:upper:2}
establishes the rates under the Margin assumption \eqref{2.12} with
$\varkappa=1$ not covered by Theorem~\ref{theoup}. In addition to
this, the classifier $\hat f_{n,2}$ of Theorem~\ref{th:upper:2} does
not require the knowledge of the margin parameter $\alpha.$ Thus,
this method is adaptive to the margin parameter. On the other hand,
the classifier $\hat f_{n,1}$ of Theorem~\ref{theoup} does require
the knowledge of $\alpha$ which is involved in the definition of
parameter $\eps$ of the net ${\cal N}_{\eps}.$ Note that for classes
${\cal F}$ of high complexity (with $\rho>1$) the empirical risk
minimization over the whole class ${\cal F}$ usually does not
provide optimal convergence rates. In such cases, some form of
regularization is needed. It could be based on penalized empirical
risk minimization (see, e.g., \cite{koltch_11}) over proper sieves
of subclasses of ${\cal F}$ (for instance, sieves of $\eps$-nets for
${\cal F}$). %Finally, we note that both classifiers $\hat f_{n,1}$
%and $\hat f_{n,2}$ have only a theoretical value since they are obtained
%as solutions of non-convex minimization problems; the discreteness of the
%minimization set for $\hat f_{n,1}$ does not help since the
%cardinality of this set is exponential in $n$.

%%%%%%%%%%%%%%%%%%%%%%%%%%%%%%%%
\section{Minimax lower bounds \label{lower}}
% %%%%
\subsection{ A general inequality}

For two probability measures $\mu$ and $\nu$ on a measurable space
$({\mathcal X}, {\mathcal A})$, we define the Kullback-Leibler
divergence and the $\chi^2$-divergence as follows:
\begin{equation}
\K(\mu,\nu) \triangleq   \int_{{\mathcal X}} g\ln g d\nu, \quad
\chi^2(\mu,\nu) \triangleq \int_{{\mathcal X}} (g-1)^2 d\nu,
\label{2.3}
\end{equation}
if $\mu$ is absolutely continuous with respect to $\nu$ with
Radon-Nikodym derivative $g=\frac{d\mu}{d\nu},$ and we set
$\K(\mu,\nu)\triangleq+\infty$, $\chi^2(\mu,\nu)\triangleq+\infty$
otherwise.

We will use the following auxiliary result.

 \begin{lemma} \label{lem2.1}
Let  $({\mathcal X}, {\mathcal A})$ be a measurable space and let
$A_i \in {\mathcal A}$, $i\in\{ 0,1,\dots,M\}$, $M\ge 2$, be such
that $\forall i\neq j$, $A_i\cap A_j =\emptyset.$ Assume that $Q_i$,
$i\in\{0,1\dots,M\}$, are probability measures on $({\mathcal X},
{\mathcal A})$ such that
$$
\frac1M\sum_{j=1}^M \K(Q_j,Q_0) \le \chi <\infty.
$$
Then
$$
p_*\triangleq \max_{0\le i \le M} Q_{i}({\mathcal X}\setminus A_i)
\ge \frac1{12}\min\{1, \, M e^{-3\chi}\}\,.
$$
\end{lemma}
\begin{Proof}
Proposition 2.3 in \cite{tsy_book} yields:
$$
p_*\ge \sup_{0<\tau<1}\frac{\tau M}{\tau M +1} \left(1-\frac{\chi +
\sqrt{\chi/2}}{\log\tau}\right).
$$
In particular, taking  $\tau^*=\min(M^{-1}, e^{-3\chi})$ and using
that $\sqrt{6\log M} \ge 2$ for $M\ge2$, we obtain
$$
p_*\ge \frac{\tau^* M}{\tau^* M +1}\left(1-\frac{\chi +
\sqrt{\chi/2}}{\log\tau^*}\right)\ge \frac1{12}\min\{1, \, M
e^{-3\chi}\}.
$$
\end{Proof}

We now prove a classification setting analogue of the lower
bound obtained by DeVore \textit{et al.} \cite{DKPT} in the
regression problem.
\begin{theorem}\label{theo2.1} Assume that a class $\Theta$ of
probability distributions $P$ with the corresponding regression functions
$\eta_P$ and Bayes rules $f^*_{P}$ (as defined above), contains a
set  $\{{P_i}\}_{i=1}^N \subset \Theta$, $N\ge 3$, with the
following properties: the marginal distribution of $X$ is $\mu_X$
for all $P_i$, independently of $i$, where $\mu_X$ is an arbitrary
probability measure, $1/4\le \eta_{P_i}\le 3/4$, $i=1,\dots,N$, and
for any $i\neq j$
\begin{equation}
\|\eta_{P_i}-\eta_{P_j}\|_{L_2(\mu_X)}\le\gamma, \label{2.5}
\end{equation}
\begin{equation}
\|f^*_{P_i}-f^*_{P_j}\|_{L_1(\mu_X)}\ge s \label{2.6}
\end{equation}
with some $\gamma>0$, $s>0$. Then for any classifier $\hat f_n$ we
have
\begin{equation}
\max_{1\le k \le N}\mathbb {P}_k\{\|\hat
f_n-f^*_{P_k}\|_{L_1(\mu_X)}\ge s/{2}\} \ge \frac1{12}\min\big(1, \,
(N-1) \exp\{-12n\gamma^2\}\big)
 \label{2.7}
\end{equation}
where $\mathbb {P}_k$ denotes the product probability measure associated to the i.i.d. $n$-sample from $P_k$.
\end{theorem}

\begin{Proof} We apply Lemma \ref{lem2.1} where we set $Q_i=\mathbb {P}_i$, $M=N-1$, and define the random events $A_i$ as follows:
$$
A_i\triangleq \{{\mathcal D}_n:\|\hat
f_n-f^*_{P_i}\|_{L_1(\mu_X)}<s/2\},\quad i=1,\dots, N.
$$
The events $A_i$ are disjoint because of \eref{2.6}. Thus, the
theorem follows from Lemma ~\ref{lem2.1} if we prove that $
\K(\mathbb {P}_i,\mathbb {P}_j)\le 4n\gamma^2$ for all $i,j$.

Let us evaluate $ \K(\mathbb {P}_i,\mathbb {P}_j)$. For each
$\eta_{P_i}$,   the corresponding measure $P_i$ is determined as
follows
\begin{equation}
dP_i(x,y)\triangleq(\eta_{P_i}(x)d\delta_{1}(y)+
(1-\eta_{P_i}(x))d\delta_{0}(y))d\mu_X(x),\label{2.8}
\end{equation}
where   $d\delta_\xi$ denotes the Dirac measure with unit mass at
$\xi$.  Set for brevity  $\eta_i\triangleq\eta_{P_i}$. Fix $i$ and
$j$.   We  have $dP_i(x,y)= g(x,y)dP_j(x,y)$, where
 $$
g(x,1)= \frac{\eta_i(x)}{\eta_j(x)},\quad g(x,0)=\frac{1-\eta_i(x)}{1-\eta_j(x)}.
 $$
Therefore, using the inequalities $1/4\le \eta_{i}, \eta_j\le 3/4$ and (\ref{2.5}) we find
\begin{align}\nonumber
\chi^2({P}_i, {P}_j)&=
\int\left\{\frac{(\eta_i(x)-\eta_j(x))^2}{\eta_j(x)}+
\frac{(\eta_i(x)-\eta_j(x))^2}{1-\eta_j(x)}\right\}d\mu_X(x)\\
&\le 8\|\eta_i-\eta_j\|_{L_2(\mu_X)}^2 \le 8\gamma^2.
\label{chi2}
\end{align}
Together with inequality between the Kullback and
$\chi^2$-divergences, cf.~\cite{tsy_book}, p.~134, this yields
$$
\K(\mathbb {P}_i,\mathbb {P}_j) = n\K({P}_i, {P}_j) \le
n\chi^2({P}_i,{P}_j)/2 \le 4n\gamma^2.
$$
\qed
\end{Proof}

%%%%
%%%%%

\subsection{Construction of a finite family of measures}
\label{sec3.2}

Theorem \ref{theo2.1} can be applied in various situations by
choosing suitable probability measures $P_i,\; i=1,\ldots, N$. In
this section, we suggest such a particular choice,  which  will give
lower bounds for classification.

Let $\s=(\s_1,\dots,\s_b)$ be a binary vector of length $b$ with
elements $\s_j\in\{-1,1\}$. Let $\fee$ be an infinitely
differentiable function with compact support in $\Rd$ such that
$0\le \fee(x)\le c$ for some constant $c\in (0,1/2)$. Consider
functions $\fee_1,\dots,\fee_b$ on $\Rd$ satisfying:

a) $\fee_j$ is a shift of $\fee$, $j=1,\dots,b$,

b) the supports $\Delta_j$ of functions $\fee_j$ are disjoint.

Denote by $\Sigma(b)$ the set of all binary vectors $\s$ of length $b$.
For every $\s\in\Sigma(b)$ define
$$
\phi_\s(x)\triangleq\sum_{j=1}^b\s_j \fee_j(x),\quad
\eta_\s(x)\triangleq (1+\phi_\s(x))/2.
$$
Consider the following class $\Theta$ of regression functions
$$
\Theta\triangleq\{\eta_\s,\s\in\Sigma(b)\}.
$$
In what follows we assume without loss of generality that $b\ge 16$.
By the Varshamov-Gilbert lemma (cf. \cite{tsy_book}, p. 104), there
is a subset $S$ of $\Sigma(b)$ such that cardinality $|S|\ge
2^{b/8}$, and for any two different elements $\s$ and $\s'$ from $S$
we have
\begin{equation}\label{low1}
\|\s-\s'\|_{\ell_1}\ge b/4.
\end{equation}
Let $\mathcal{X}=[0,1]^d$, $q\in \Bbb N$, and $b\triangleq q^d$. Let
$\psi$ be a nonnegative infinitely differentiable
function with support $(0,1)^d$ such that $\psi\le c<1/2$ and
$\int_{(0,1)^d}\psi(x)dx > 0$. For given parameters $\delta\in
(0,1)$ (small parameter) and $\alpha \in [0,\infty),$ define
$$
\fee(x)\triangleq\dt^{1/(1+\alpha)}\psi(qx).
$$
For a vector $k=(k_1,\dots,k_d)$, $k_j\in \{0,\dots, q-1\}$,
$j=1,\dots,d,$ define a grid point
$$
x^k\triangleq (x_1^k,\dots,x_d^k), \quad  x_j^k= k_j/q,\
j=1,\dots,d.
$$
We now consider $b$ functions $\fee_k(x)=\fee(x-x^k)$ and the
corresponding class $\Theta$ of regression functions defined above.
We set $N\triangleq|S|$ and consider a subset $\Theta'\subset
\Theta$:
$$
\Theta'\triangleq\{\eta_\s,\s\in S\} = \{\eta_{i}\}_{i=1}^N.
$$
Now, recalling that the regression function $\eta(X)$ is the
conditional probability of $Y=1$ given $X$, we define the joint
probability measures $P_\s,\s\in S,$ of $(X,Y)$ (these measures will
be also denoted by $P_i, i=1,\dots, N$) :
$$
P_\s (Y=1, X\in A) = \int_A\eta_\s(x) \mu_X(dx)%,\quad {\rm with} \
$$
for any Borel set $A$, where the marginal distribution
$\mu_X=\mu_X^*$ is specified as follows. First, for all $x$ such
that
$$
1/(4q)\le x_j-x^k_j\le 3/(4q),\quad j=1,\dots,d,
$$
the distribution $\mu_X^*$ has a density w.r.t. the Lebesgue
measure
\begin{equation}\label{eq:construc}
\frac{d\mu_X^*}{dx}(x)\triangleq \frac{w}{{\rm Leb}(B(0,1/(4q))}=2^dbw
\end{equation}
where $B(x,r)$ is the $\ell_\infty$-ball of radius $r$ centered at
$x$, ${\rm Leb}(\cdot)$ denotes the Lebesgue measure, and
$w=C\dt^{\alpha/(1+\alpha)}/b$ for some $C\in (0,1]$. Second, we set
${d\mu_X^*}(x)/{dx}=0$ for all other $x$ such that at least one of
$\eta_{i}(x)$ is not $1/2$. Finally, on the complementary set
$A_0\subset [0,1]^d$ where all $\eta_{i}(x)$ are equal to $1/2$, we
set $d\mu_X^*(x)/dx\triangleq(1-bw)/{\rm Leb}(A_0)$ to ensure that
$\int_{\Rd}d\mu_X^*(x)=1$ (we assume that the support of the function $\psi$
belongs to the set $[\gamma, 1-\gamma]$ for a small $\gamma>0;$ then, it is easy
to see that ${\rm Leb}(A_0)>0$).

We now impose an extra restriction on $\fee$ and prove that under
this restriction the measures $P_i$ satisfy the Margin condition
with parameter $\alpha$. Assume that $\psi(x)=c_2>0$ for $x$
satisfying the inequalities $1/4\le x_j \le 3/4$, $j=1,\dots,d$, and
$\psi(x)<c_2$ for other $x$. Here $c_2\in (0,1/2)$. Then
\begin{align*}
\mu_X^*(0<|\eta_\s(X)-1/2|\le t) &=
\mu_X^*(0<|\sum_{j=1}^b\s_j\fee_j(X)|\le 2t) \\
&= b\mu_X^*(0<\fee(X)\le 2t),
\end{align*}
because the supports $\Delta_j$ of functions $\fee_j$ are disjoint.
Then, using the definition
$\fee(x)\triangleq\dt^{1/(1+\alpha)}\psi(qx)$ we obtain that
$$
\mu_X^*(0<\fee(X)\le 2t) = w \quad\text{if}\quad
c_2\dt^{1/(1+\alpha)}\le 2t
$$
and $\mu_X^*(0<\fee(X)\le 2t)=0$ for all other $t>0$. Therefore,
$$
b\mu_X^*(0<\fee(X)\le 2t)\le
C\dt^{\alpha/(1+\alpha)}{{I}}_{\{c_2\dt^{1/(1+\alpha)}\le 2t\}}\le
C(2t/c_2)^\alpha, \quad t>0.
$$
Thus, all $P_i$ satisfy the Margin condition with parameter $\alpha$
and constant $C_M=C(2/c_2)^\alpha$.

\subsection{Minimax lower bound for classification}

Let us check the assumptions of Theorem \ref{theo2.1} for the set of
probability measures $P_1,\dots, P_N$ defined in
Section~\ref{sec3.2}. Since $0<c<1/2$ we have $1/4\le \eta_{i}(x)\le
3/4$ for all $\dt\in(0,1)$ and all $x\in(0,1)^d$. Next, for any
$\s,\s'\in S$ we have
\begin{equation}
\|\eta_{P_\s}-\eta_{P_{\s'}}\|_{L_2(\mu_X^*)}^2 \le
b\|\fee\|_{L_\infty(\mu_X^*)}^2 w \le C \dt^{(2+\alpha)/(1+\alpha)},
\label{3.2}
\end{equation}
and for $\sigma\neq \sigma'$, in view of (\ref{low1}) and
(\ref{eq:construc}),
\begin{eqnarray}
\nonumber \|f^*_{P_\s}-f^*_{P_{\s'}}\|_{L_1(\mu_X^*)} &=&
2\sum_{j=1}^b {{I}}_{\{\s_j\ne\s'_j\}}\int_{B(0,1/(4q))}2^dbw \, dx
\\
&=&\|\s-\s'\|_{\ell_1}w \ge c_1 \dt^{\alpha/(1+\alpha)}, \label{3.3}
\end{eqnarray}
where $c_1=C/4$. Thus, the assumptions of Theorem \ref{theo2.1} are
satisfied with $N=|S|\ge 2^{b/8}\ge 2^{b/16} +1$, and
\begin{equation}
\gamma^2 =C \dt^{(2+\alpha)/(1+\alpha)},\qquad s = c_1
\dt^{\alpha/(1+\alpha)}.\label{3.4}
\end{equation}
Therefore, we get the following result.
\begin{proposition}\label{prop:lower} Fix $\alpha>0$, $\delta\in
(0,1)$ and $q\in \Bbb N$ such that $b=q^d\ge 16$. Let $P_1,\dots,
P_N$ be the family of probability measures defined in Section
\ref{sec3.2}. Then for any classifier $\hat f_n$ we have
\begin{equation}
\max_{1\le k \le N}\mathbb{P}_k\Big\{ \|\hat
f_n-f^*_{P_k}\|_{L_1(\mu_X^*)}\ge
\frac{C\dt^{\frac{\alpha}{1+\alpha}}}{8} \Big\} \ge
\frac1{12}\min(1,
2^{\frac{b}{16}}\exp\{-c_3n\dt^{\frac{2+\alpha}{1+\alpha}}\})
\label{3.5}
\end{equation}
where $C\in (0,1)$ is the constant used in the construction of
Section \ref{sec3.2}, and $c_3>0$ is a constant depending only on
$C$. Furthermore, for $0<\lambda <\lambda_0$,
\begin{align}\label{3.5a}
\max_{1\le k \le N}\mathbb{P}_k\{ R(\hat f_n)-R(f^*_{P_k}) \ge
\lambda\} & \ge \frac1{12}\min(1,
2^{\frac{b}{16}}\exp\{-c_4n\lambda^{\frac{2+\alpha}{1+\alpha}}\})
\end{align}
where $\lambda_0=16^{-(1+\alpha)/\alpha}Cc_2$, and $c_4>0$ is a
constant depending only on $C$, $c_2$ and~$\alpha$.
\end{proposition}
\begin{Proof} Bound (\ref{3.5}) follows from Theorem
\ref{theo2.1} and (\ref{3.4}). To prove (\ref{3.5a}), we combine
(\ref{3.5}) with Lemma~\ref{lem2.3}, set $\lambda=\lambda_0\delta$,
and use that $C_M=C(2/c_2)^\alpha$ by the construction of Section
\ref{sec3.2}.
\end{Proof}

\subsection{Application to a particular class of distributions}
\label{sec:lower:holder}

In this section, we will assume that the regression function $\eta$
belongs to a H\"older class defined as follows.

For any multi-index $s=(s_1,\dots,s_d)$ and any
$x=(x_1,\dots,x_d)\in{\mathbb R}^d$, we define $|s| = \sum_{i=1}^d
s_i$, $s!=s_1!\dots s_d!$, $x^s=x_1^{s_1} \dots x_d^{s_d}$ and
$\|x\| \triangleq (x_1^2+\cdots+x_d^2)^{1/2}$. Let $D^s$ denote the
differential operator
    $D^s\triangleq \frac{\partial^{s_1+ \cdots +s_d}}{\partial x_1^{s_1}
        \cdots \partial x_d^{s_d}}.$

For $\beta >0$, let $\lfloor \beta\rfloor $ be the maximal integer
that is strictly less than $\beta$. For any $x\in[0,1]^d$ and any
$\lfloor \beta\rfloor$ times continuously differentiable real valued
function $g$ on $[0,1]^d$, we denote by $g_x$ its Taylor polynomial
of degree $\lfloor \beta\rfloor$ at point $x\in[0,1]^d$:
    $$g_x(x') \triangleq \sum_{|s|\le \lfloor \beta\rfloor}
    \frac{(x'-x)^s}{s!} D^s g(x).$$

Let $\beta >0$, $L>0$. The {\it H\"older class} of functions
$\Sigma( \beta , L , [0,1]^d )$ is defined as the set of all
functions $g:[0,1]^d\to \R$ that are $\lfloor \beta\rfloor$ times
continuously differentiable and satisfy, for any $x,y\in [0,1]^d$ ,
the inequality
        $$|g(x') - g_x(x')| \le L \|x'-x\|^{\beta}.$$

We now apply the technique of proving minimax lower bounds developed
in the previous sections to the following class of distributions.

Fix $\alpha>0, \beta>0, L>0$, and a probability distribution $\mu_X$
on $[0,1]^d$. Denote by $M'(\mu_X,\alpha,\beta)$ the class of all
joint distributions $P$ of $(X,Y)$ such that:  {\it
\begin{itemize}
\item[(i)] The marginal distribution of $X$ is $\mu_X$;
\item[(ii)] The Margin condition \eref{2.15} is satisfied with some constant
$C_M>0$;
\item[(iii)] The regression function $\eta=\eta_P$ belongs to the H\"older class $\Sigma(\beta, L, [0,1]^d)$.
\end{itemize}
}
%Consider the union of such classes:
%\begin{align}\displaystyle{{\mathcal M'}(\alpha,\beta)=\cup_{\mu_X\in {\mathcal B}}M'(\mu_X,\alpha,\beta)}\label{def:M'}
%\end{align}
%where ${\mathcal B}$ denotes the class of all probability
%distributions on $[0,1]^d$.

%%%%%%%%%%%%%%%%%%
\begin{theorem}\label{th:lowerbound}
 Let $\mu_X^*$ be the marginal density defined in Section~\ref{sec3.2}.   There exist positive constants $C'_1,\;C'_2,\; c'$ and $d'_1, d'_2, \lambda_0'$ depending only on $ \alpha, \beta, L, d$, and $C_M$ such that for any classifier $\hat f_n$,
$$\sup_{P\in {\mathcal M'}( \mu_X^*,\alpha,\beta)} \mathbb{P}\{
R(\hat f_n)-R^*\ge \lambda\}\ge C'_1
$$
for any $0<\lambda\le d'_1 n^{-\frac{1+\alpha}{2+\alpha+d/\beta}}$,
and
$$\sup_{P\in {\mathcal M'}(\mu_X^*,\alpha,\beta)} \mathbb{P}\{
R(\hat f_n)-R^*\ge \lambda\}\ge C'_2\exp\{-c'n
\lambda^{\frac{2+\alpha}{1+\alpha}} \}
$$
for any $ d'_2 n^{-\frac{1+\alpha}{2+\alpha+d/\beta}}\le \lambda \le \lambda_0'$.
\end{theorem}
%%%%%%%%%%%%%%%%
\begin{Proof} Set
$q=\lceil c_5\delta^{-\frac1{(1+\alpha)\beta}}\rceil$ where $c_5>0$
is a constant, and $\lceil x \rceil$ denotes the minimal integer
greater than $x$. It is easy to see that if $c_5$ is small enough,
then (see Section~\ref{sec3.2}) we have $\fee \in \Sigma(\beta,
L,[0,1]^d)$ implying that $\eta_\sigma \in \Sigma(\beta, L,[0,1]^d)$
for all $\sigma\in S$. Choose such a small $c_5$. It is also
easy to see that one can always choose constants $C\in (0,1)$ and
$c_2\in (0,1/2)$ in the construction of Section~\ref{sec3.2} in such a
way that $C(2/c_2)^{\alpha}\leq C_M$ which is needed to satisfy the
margin condition (ii).
Then, for any
fixed $\delta \in (0,1)$, the finite family of probability
distributions $\{P_1,\dots,P_N\}$ constructed in
Section~\ref{sec3.2} and depending on $\delta$ belongs to ${\mathcal
M'}(\mu_X^*,\alpha,\beta)$. To indicate this dependence on $\delta$
explicitly, denote this family by ${\mathcal P}_\lambda$ where
$\lambda=\lambda_0\delta$ and $\lambda_0$ is defined in
Proposition~\ref{prop:lower}. Since ${\mathcal P}_\lambda\subset
{\mathcal M'}(\mu_X^*,\alpha,\beta)$, for any $\lambda<\lambda_0$ we
can write
$$
\sup_{P\in {\mathcal M'}(\mu_X^*, \alpha,\beta)} \mathbb{P}\{ R(\hat
f_n)-R^*\ge \lambda\}\ge \max_{P\in {\mathcal P}_\lambda}
\mathbb{P}\{ R(\hat f_n)-R^*\ge \lambda\}
$$
and then estimate the right hand side of this inequality using
(\ref{3.5a}) of Proposition~\ref{prop:lower}. Note that in
Proposition~\ref{prop:lower} we have the assumption $q^d\ge 16$,
which is satisfied if $\delta\le \delta_0$ where $\delta_0$ is a
small enough constant depending only on the constants in the
definition of the class ${\mathcal M'}(\mu_X^*, \alpha,\beta)$. Thus
we obtain
\begin{eqnarray*}
\sup_{P\in {\mathcal M'}( \mu_X^*,\alpha,\beta)} \mathbb{P}\{ R(\hat
f_n)-R^*\ge \lambda\}&\ge& \frac1{12}\min(1, 2^{b/16}
\exp\{-c_4n\lambda^{\frac{2+\alpha}{1+\alpha}}\})\\
& \ge& \frac1{12}\min(1,
\exp\{c_6\lambda^{-\frac{d}{(1+\alpha)\beta}}
-c_4n\lambda^{\frac{2+\alpha}{1+\alpha}}\})
\end{eqnarray*}
for all $0<\lambda<\lambda_0'$ where $\lambda_0'>0$ and $c_6>0$
depend only on the constants in the definition of the class
${\mathcal M'}(\mu_X^*, \alpha,\beta)$. This immediately implies the
theorem. \qed
\end{Proof}

%%%%%%%%%%%%%%%%
%The results above can be expressed in terms of AC-functions.
%\begin{corollary}
% There exist   $\lambda_n^-\asymp \lambda_n^+\asymp n^{-\frac{1+\alpha}{2+\alpha+r\alpha}}$, and %constants $\delta_0,\; C',\; c'$ such that
%\begin{align}\label{quote11}
%AC_n({\mathcal M'}(r , \alpha) ,\lambda)&\ge \delta_0,\qquad \forall\; \lambda\le \lambda_n^-,\\
%AC_n({\mathcal M'}(r , \alpha) ,\lambda)&\ge C'
%e^{-c'n\lambda^{\frac{2+\alpha}{1+\alpha}}},\qquad \forall\;
%\lambda\ge \lambda_n^+. \label{quote21}
%\end{align}
%\end{corollary}
%%%%%%%%%%%%%%%%

\medskip

Note that the class of distributions $M'(\mu_X^*,\alpha,\beta)$ has
the following properties.
\begin{itemize}
%\item[(I)] The Margin condition \epsref{2.15} holds for any
%$P\in {\mathcal M'}(\alpha,\beta)$;
\item[(A)] There exists a constant $B>0$ such that the set of regression functions ${\mathcal
U}=\{\eta_P$, $P\in \mathcal M'(\mu_X^*,\alpha,\beta)\}$ satisfies the
entropy bound
\begin{equation}
\mathcal H(\eps,{\mathcal U},\|\cdot\|_{\infty})\le B\eps^{-r}, \
\forall \eps>0, \label{3.7a} \end{equation} where $r=d/\beta$.
\item[(B)]  There exists a constant $B'>0$ such that the set of Bayes classifiers $\mathcal
F=\{f^*_P$, $P\in \mathcal M'(\mu_X^*,\alpha,\beta)\}$ satisfies the
bracketing entropy bound
\begin{equation}
 \mathcal H_{[\ ]}(\eps,\mathcal F,\|\cdot\|_{L_1(\mu_X^*)})\le
B'\eps^{-\rho}, \ \forall \eps>0, \label{3.7} \end{equation} where
$\rho= d/(\alpha\beta)$. \end{itemize} Indeed, (A) holds since
$\mathcal U=\{\eta\in\Sigma(\beta,L,[0,1]^d): \,0\le \eta(x)\le
1\}$, and
$$\mathcal H(\eps,\Sigma(\beta,L,[0,1]^d),\|\cdot\|_{\infty})\le B\eps^{-d/\beta},
$$
cf. Kolmogorov and Tikhomirov~\cite{kt61}. Moreover, this bound
holds if we replace the $\eps$-entropy $\mathcal
H(\cdot,\cdot,\cdot)$  by the bracketing $\eps$-entropy $\mathcal
H_{[\ ]}(\cdot,\cdot,\cdot)$ depending on the same arguments, cf.
Dudley \cite{Dudley}. This and Corollary~\ref{cor1}
imply~(\ref{3.7}).

In conclusion, for the choice of $\mu_X^*$ described in Section
\ref{sec3.2}, the class of probability distributions $\mathcal
M'(\mu_X^*,\alpha,\beta)$ is a particular case of both $\mathcal
M(r,\alpha)$ (with $r=d/\beta$) and of $\mathcal M^*(\rho,\alpha)$
(with $\rho= d/(\alpha\beta)$ and $\mu=\mu_X^*$) defined in
Sections~\ref{sec:upper:reg} and~\ref{sec:upper:Bayes}.
Theorem~\ref{th:lowerbound} shows that, for this particular case, it
is impossible to obtain faster rates for $AC$-functions than those
established in Theorems~\ref{theoup} and~\ref{th:upper:2}. In this
sense, Theorem~\ref{th:lowerbound} provides a lower bound that
matches the upper bounds of Theorems~\ref{theoup}
and~\ref{th:upper:2}.

\footnotesize{

}

\end{document}